\documentclass[journal,10pt,final]{IEEEtran}

\usepackage{graphicx}
\usepackage{epstopdf}
\usepackage{amsthm,amsmath,amssymb}
\usepackage{mathrsfs}
\usepackage{setspace}
\usepackage{amsmath}
\usepackage{booktabs}
\usepackage{enumerate}
\usepackage{cite}
\usepackage{stfloats}
\usepackage{subfigure}

\usepackage{algorithm}
\usepackage{algorithmicx}
\usepackage{algpseudocode} 
\usepackage{multirow}
\usepackage{flafter}
\usepackage{bm}

\usepackage{url}
\usepackage{xcolor}
\usepackage{subfigure}
\usepackage{fancyhdr}
\usepackage{mdwmath}
\usepackage{mdwtab}
\usepackage{caption}
\usepackage{amsthm}
\usepackage{setspace}
\usepackage{tabularx}
\usepackage{tabu}
\usepackage{lipsum}

\newtheorem{remark}{Remark}

\usepackage[
top    = 2.71cm,
bottom = 1.05in,
left   = 0.6 in,
right  = 0.6 in]{geometry}

\allowdisplaybreaks[4]

\ifCLASSINFOpdf
\else
\fi

\begin{document}

\title{SLARM: Simultaneous Localization and Radio Mapping for Communication-aware Connected Robot}

\author{\IEEEauthorblockN{ Xinyu~Gao\IEEEauthorrefmark{1}, Yuanwei~Liu\IEEEauthorrefmark{1}, and Xidong~Mu\IEEEauthorrefmark{2}} \\ 
\IEEEauthorblockA{\IEEEauthorrefmark{1} Queen Mary University of London, London, UK\\
\IEEEauthorrefmark{2} Beijing University of Posts and Telecommunications, Beijing, China\\\vspace{-0.3cm}
 }}

\maketitle

\begin{abstract}
  A novel simultaneous localization and radio mapping (SLARM) framework for communication-aware connected robots in the unknown indoor environment is proposed, where the simultaneous localization and mapping (SLAM) algorithm and the global geographic map recovery (GGMR) algorithm are leveraged to simultaneously construct a geographic map and a radio map named a channel power gain map. Specifically, the geographic map contains the information of a precise layout of obstacles and passable regions, and the radio map characterizes the position-dependent maximum expected channel power gain between the access point and the connected robot. Numerical results show that: 1) The pre-defined resolution in the SLAM algorithm and the proposed GGMR algorithm significantly affect the accuracy of the constructed radio map; and 2) The accuracy of radio map constructed by the SLARM framework is more than 78.78\% when the resolution value smaller than 0.15m, and the accuracy reaches 91.95\% when the resolution value is pre-defined as 0.05m.
\end{abstract}

\IEEEpeerreviewmaketitle

\vspace{-0.55cm}

\section{Introduction}
In the wake of the advancement of computer technology and mechanical crafts, the mobile robots equipped with actuators and sensors with controllers turn into automatic mechanism apparatus progressively \cite{IEEEhowto:Latombe}. Robots can befriend humans to implement miscellaneous even dangerous works. However, the video data or the 3D point cloud data is inevitably pulled into when the robot using the equipped computation resources to handle assigned special missions. And the consumption of a large number of computing resources renders the sky-high cost of the hardware when robots are applied for repeatedly handling high dimensional data locally. Accordingly, the introduction of a cloud control center can assist robots to compute the results for policy-making, namely, communication-aware connected robot, to support reducing the complexity of local calculations. Additionally, in the past few years, the explosive growth of user access throughput and data throughput has spawned fifth-generation (5G) and beyond 5G (B5G) networks \cite{IEEEhowto:Andrews}. The high speed, high reliability and low latency of the 5G network enable the creation of the shared link between robots and access points (APs). Thus, 5G cellular networks aided autonomous robots can fulfill the demand for efficient work.
\par
Localization and geographic map constructions are the essential components for robot navigation for accomplishing various tasks\cite{IEEEhowto:Gao}. In pace with the maturity of various sensing technologies, robot localization studies becomes appealing. The authors in \cite{IEEEhowto:QZhang} proposed an algorithm runs in time linear in the number of landmarks by making efficient use of the representation of the landmarks by complex numbers, which is described as an efficient method for localizing a mobile robot in an environment with landmarks. An indoor inertial navigation system (INS) integrated with light detection and ranging (LiDAR) robot localization system is proposed in \cite{IEEEhowto:YXu} to provide accurate information about the robot location. In \cite{IEEEhowto:LMarinho}, a novel approach for mobile robot localization from images is proposed based on supervised learning using topological representations for the environment, where the spatial Moments combined with Bayes classifier is the best performing model, providing high accuracy rate and small computational time. The utilization of internet of things (IoT) for the development of a system aimed for localization mobile robots employing convolutional neural networks (CNN) in the process of feature extraction of the images, according to the concept of transfer learning is evidenced in \cite{IEEEhowto:Junior}. In \cite{IEEEhowto:Zhao}, a fault-tolerance architecture is proposed for mobile robot localization and a differential drive mobile robot is investigated. The effectiveness of the fault-tolerance architecture is verified in several experiments conducted in the robot prototype. An indoor robot VLP localization system based on Robot Operating System (ROS) is presented in \cite{IEEEhowto:Guan} for the first time, aiming at promoting the application of VLP in mature robotic system. The authors in \cite{IEEEhowto:GLi} presented an improved observation model for Monte-Carlo localization (MCL), which improves the robustness of localization by reliable reflector prediction in the ambiguous environments caused by incorrect reflectors detection. 
\par
While the aforementioned research contributions have laid a solid foundation on robot localization, the investigations on the localizations for communication-aware connected robot are still quite open, especially in unknown environment. The limitations and challenges are summarized as follows: 1) For the communication-aware connected robot, the propagation channel between the AP and the connected  robot can become weak when the communication link is blocked by the obstacles. The resulting position-dependent channel model challenges the application of the connected robots since they are communication-sensitive. 2) For the unknown environment, the move security for robots should be well guaranteed.
\par
In response to the above limitations and challenges, our study draws on the SLAM algorithm and extends to simultaneous localization and radio mapping (SLARM) framework. In this framework, a geographic map and a radio map can be constructed simultaneously, which both ensure the move security and communication quality of the connected robots. Specifically, a global geographic map recovery (GGMR) algorithm is developed for determining the geographic map, which contains the information of a precise layout of obstacles and passable regions. Based on the geographic map, a radio map is simultaneously constructed to characterizes the spatial distribution of the maximum expected channel power gain between the AP and the robot. Numerical results show that the resolutions of the SLARM algorithm plays a significant role in possible communication area utilization.

\vspace{-0.5cm}
\section{System Model}
\vspace{-0.2cm}
\setlength{\belowcaptionskip}{-0.3cm}
\begin{figure}[ht] 
  \centering  
  \includegraphics[height=1.4in,width=2.8in]{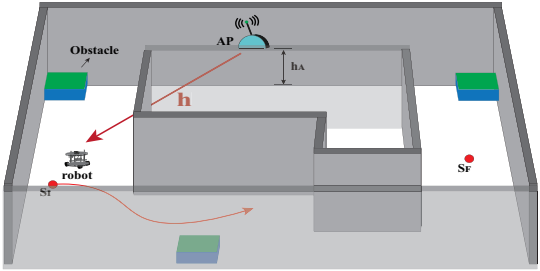}
  \caption{Illustration of the communication-aware connected robot system in an unknown indoor environment.}
  \label{IRS}
\end{figure}
In this paper, we consider a communication-aware connected robot system in an unknown indoor environment. As shown in Fig.~\ref{IRS}, the environment consists of one single-antenna AP and one single-antenna mobile robot. Without loss of generality, we assume that the floor is smooth with slight undulation, whose impact led to the system can be negligible. When a three-dimensional (3D) Cartesian coordinate system is established based on the plane of the ground, the mobile space and the positions of AP are specified by [$-\emph{x}_{max}^{\pmb{M}},\emph{x}_{max}^{\pmb{M}}$] $\times$ [$-\emph{y}_{max}^{\pmb{M}},\emph{y}_{max}^{\pmb{M}}$] and \emph{$(0^{\pmb{M}},y_{max}^{\pmb{M}},h_{1})$}. \emph{$\pmb{M} \in \mathbb{R}^{X \times Y}$} denotes the passable move space by the robot exploration. Among them, $\emph{x}_{max}^{\pmb{M}}$ and $\emph{y}_{max}^{\pmb{M}}$ denote the maximum bound of the \emph{x} and \emph{y} coordinate of the moving space, respectively. Additionally, let \emph{$h_{m}$} denote the vertical height of the received antenna equipped on the robot, and mark \emph{$S^{\pmb{M}}=(x^{\pmb{M}},y^{\pmb{M}},h_{m})$} denote the position of the robot. The equivalent baseband Rician channel between the AP and the robot at position \emph{$S^{\pmb{M}}$} can be expressed as
\vspace{-0.2cm}
\begin{align}\label{1}  
  \pmb{h}(S^{\pmb{M}}) = \sqrt{\frac{\alpha_{s}(S^{\pmb{M}})}{\alpha_{s}(S^{\pmb{M}})+1}}\tilde{\pmb{h}}+\sqrt{\frac{1}{\alpha_{s}(S^{\pmb{M}})+1}}\hat{\pmb{h}},
\end{align}
\vspace{-0.3cm}
\par
\noindent
where \emph{$\alpha_{s}(S^{\pmb{M}})$}, \emph{$\tilde{\pmb{h}}$} and \emph{$\hat{\pmb{h}}$} denote the position-dependent Rician factor, position-dependent LoS components, and position-dependent LoS components, respectively. Specifically, if the the communication link between the AP and the robot at the location \emph{$S^{\pmb{M}}$} is blocked, \emph{$\alpha_{s}(S^{\pmb{M}})=0$}; otherwise, \emph{$\alpha_{s}(S^{\pmb{M}})=\overline{\alpha}$}, which is a constant value. Note that the \emph{$\pmb{h}(S^{\pmb{M}})$} is a random variable, and it is position-dependent with respect to \emph{$S^{\pmb{M}}$}. Thus, we consider the expected channel power gain denoted by \emph{$\mathbb{E}[|\pmb{h}(S^{\pmb{M}})|^{2}]$}, which can be expressed as
\vspace{-0.2cm}
\begin{align}\label{1.1}
  &\mathbb{E}[{|\pmb{H}(S^{\pmb{M}})|^{2}}] = \mathbb{E}[{|\pmb{L}(S^{\pmb{M}})\pmb{h}(S^{\pmb{M}})|^{2}}]=\mathbb{E}[|\pmb{L}(S^{\pmb{M}})|^{2}] \cdot \nonumber \\
  &(\frac{\alpha_{s}(S^{\pmb{M}})}{\alpha_{s}(S^{\pmb{M}})+1}\mathbb{E}[|\tilde{\pmb{h}}|^{2}]+\frac{1}{\alpha_{s}(S^{\pmb{M}})+1}\mathbb{E}[|\hat{\pmb{h}}|^{2}])=\mathbb{E}[\pmb{L}(S^{\pmb{M}})].
\end{align}
\vspace{-0.4cm}
\par
\noindent
where the \emph{$\pmb{L}(S^{\pmb{M}})$} denotes the distance-dependent path loss between the AP and the robot. In this paper, our goal is to identify the passable region \emph{$\pmb{M}$}, which avoids collisions, and the position-dependent channel power gain. These regions help prevent connected robots from being out of control. To handle this task, a novel framework, namely simultaneous localization and radio mapping (SLARM), is proposed.

\vspace{-0.5cm}
\section{SLARM: Simultaneous Localization and Radio Mapping Framework in the unknown environment}
In this section, we will introduce the SLARM framework. The SLARM framework integrates the geographic map construction and radio map construction algorithms, which ensures to simultaneously build both two maps for communication-aware connected robots in the unknown environment.

\vspace{-0.5cm}
\subsection{Geographic map construction algorithm}
For ease of exposition, the two-dimensional (2D) space of the X-Y plane of the geographic map for the robot is discretized as \emph{$\frac{4x_{max}y_{max}}{\delta^{2}}$} grids at first. The \emph{$\delta$} denotes the discrete resolution for space, which is small enough to make the size of the grid approximate to the center point of the grid. Thus, the center of \emph{(a,b)}-th grid can be expressed as
\vspace{-0.2cm}
\begin{align}\label{9}
  S_{a,b} = &S_{I} + [a-1, b-1] \delta, \nonumber \\
  &a \in \{1,2,\cdots,\frac{2x_{max}}{\delta}\}, b \in \{1,2,\cdots,\frac{2y_{max}}{\delta}\}.
\end{align}
\vspace{-0.5cm}

\subsubsection{Simultaneous localization and mapping algorithm for sub-maps construction}
Dissimilar with performing tasks on a precise map, the robot is confronted with an unaccounted-for environment. Thus, laser-based simultaneous localization and mapping (L-SLAM) algorithms can be leveraged for geographic map founded, which is based on an optimized way of particle filtering. Considering the gmapping scheme \cite{IEEEhowto:Grisetti}, gmapping improves the proposed distribution and selective resampling according to Rao-Blackwellized Particle Filters (RBPF) \cite{IEEEhowto:Carlone}, thereby reducing the number of particles and preventing particle degradation. In the RBPF, the problem to be resolved is to seek the joint distribution of pose and map \emph{$p(x_{1:t},\hat{m}|o_{1:t},l_{1:t})$}, while the sensor data \emph{$l_{1:t}$} and observation data \emph{$o_{1:t}$} are obtained. To make the problem briefless, the conditional probability can be disassembled as position probability estimation and map drawing. Thus, it can be written as
\vspace{-0.2cm}
\begin{align}\label{10}
  p(x_{1:t},\hat{m}|o_{1:t},l_{1:t}) = p(\hat{m}|x_{1:t},o_{1:t})\cdot p(x_{1:t}|o_{1:t},l_{1:t-1}),
\end{align}
\vspace{-0.5cm}
\par
\noindent
where the \emph{$x_{1:t}$} and \emph{$\hat{m}$} denote the robot's positions and sub-maps from 1 to \emph{t}, respectively. It is worth noting that the positions of the robot in equation \eqref{10} are estimation by importance sampling algorithm (ISA). According to the ISA, after the robot’s positions at the moment \emph{t} are predicted, the state value can be sampled while the value is obtained. Next, the weight for each particle and the iterative process for the weight can be calculated as
\vspace{-0.2cm}
\begin{align}\label{11}
  &w_{t}^{(i)} = \frac{p(x_{1:t}^{(i)}|o_{1:t},l_{1:t-1})}{q(x_{1:t}^{(i)}|o_{1:t},l_{1:t-1})} \nonumber \\
  & = \frac{\eta p(o_{t}|x_{1:t}^{(i)},o_{1:t-1})p(x_{t}^{(i)}|x_{t-1}^{(i)},l_{t-1})}{q(x_{t}^{(i)}|x_{1:t-1}^{(i)},o_{1:t},l_{1:t-1})} \nonumber \\
  &\cdot \underbrace{\frac{p(x_{1:t-1}^{(i)}|z_{1:t-1},l_{1:t-2})}{q((x_{1:t-1}^{(i)}|z_{1:t-1},l_{1:t-2}))}}_{w_{t-1}^{(i)}} \nonumber \\
  & \propto \frac{p(o_{t}|\hat{m}_{t-1}^{(i)},x_{t}^{(i)})p(x_{t}^{(i)}|x_{t-1}^{(i)},l_{t-1})}{q(x_{t}^{(i)}|x_{1:t-1}^{(i)},o_{1:t},l_{1:t-1})}\cdot {w_{t-1}^{(i)}},
\end{align}
\vspace{-0.4cm}
\par
\noindent
where the \emph{$w_{t}^{(i)}$} and \emph{$\eta$} denote the weight at time \emph{t} of the \emph{i}-th iteration and scale factor, respectively. \emph{$q(x_{t}^{(i)}|x_{1:t-1}^{(i)},o_{1:t},l_{1:t-1})$} is the key indicator which assists weight determination. After the robot carries out resampling, particles can be redistributed according to the particle weights obtained by the sampling results, which provides input values for the next prediction. If the particles are directly sampled from the target distribution, merely one particle is needed to obtain the position estimation of the robot. If sampling from the sensor, the equation \eqref{11} can be rewritten as
\vspace{-0.2cm}
\begin{align}\label{12}
  w_{t}^{(i)} &= w_{t-1}^{(i)} \cdot \frac{p(o_{t}|\hat{m}_{t-1}^{(i)},x_{t}^{(i)})p(x_{t}^{(i)}|x_{t-1}^{(i)},l_{t-1})}{p(x_{t}^{(i)}|x_{t-1}^{(i)},l_{t-1})} \nonumber \\
  &\propto w_{t-1}^{(i)} \cdot p(o_{t}|\hat{m}_{t-1}^{(i)},x_{t}^{(i)}).
\end{align}
\vspace{-0.4cm}
\par
In this model, the number of particles to simulate the state distribution are with limitation. It is essential to discard the particles with low weight and let the particles with large weight replicate to achieve the convergence of the particles to the real state. However, particle degradation will be exposed while frequent re-sampling is implemented. Therefore, the further improved proposed distribution can be expressed as
\vspace{-0.2cm}
\begin{align}\label{13}
  p(x_{t}|\hat{m}_{t-1}^{(i)},x_{t_1}^{(i)},o_{t},l_{t-1}) = \frac{p(o_{t}|m_{t-1}^{(i)},x_{t})p(x_{t}|x_{t-1}^{(i)},l_{t-1})}{p(o_{t}|\hat{m}_{t-1}^{(i)},x_{t_1}^{(i)},l_{t-1})},
\end{align}
\vspace{-0.4cm}
\par
\noindent
and then particle weight can be recalculated as
\vspace{-0.2cm}
\begin{align}\label{14}
  w_{t}^{(i)} & = w_{t-1}^{(i)} \cdot \frac{\eta p(o_{t}|\hat{m}_{t-1}^{(i)},x_{t})p(x_{t}|x_{t-1}^{(i)},l_{t-1})}{p(x_{t}|\hat{m}_{t-1}^{(i)},x_{t_1}^{(i)},o_{t},l_{t-1})} \nonumber \\
  & \propto w_{t-1}^{(i)} \cdot \frac{p(o_{t}|\hat{m}_{t-1}^{(i)},x_{t}^{(i)})p(x_{t}|x_{t-1}^{(i)},l_{t-1})}{\frac{p(o_{t}|\hat{m}_{t-1}^{(i)},x_{t}^{(i)})p(x_{t}|x_{t-1}^{(i)},l_{t-1})}{p(o_{t}|\hat{m}_{t-1}^{(i)},x_{t_1}^{(i)},l_{t-1})}} \nonumber \\
  & = w_{t-1}^{(i)} \cdot p(o_{t}|\hat{m}_{t-1}^{(i)},x_{t_1}^{(i)},l_{t-1}) \nonumber \\
  & = w_{t-1}^{(i)} \cdot \int p(o_{t}|x')p(x'|x_{t-1}^{(i)},l_{t-1})dx'.
\end{align}
\vspace{-0.4cm}
\par
The sampling method can be used to simulate the proposed distribution since an approximate form of accurate target distribution is unavailable. There is a few peaks but one peak in most case in the target distribution, while it can be sampled from the peak directly. Thus, the Gaussian function can be simulated as a proposed distribution after \emph{z} values are selected near the peak. The Mean and variance can be calculated as
\vspace{-0.2cm}
\begin{align}\label{15}
  \mu_{t}^{(i)} = \frac{1}{\eta^{(i)}} \cdot \sum\limits_{j=1}^{z} x_{j} \cdot p(o_{t}|\hat{m}_{t-1}^{(i)},x_{j}) \cdot p(x_{j}|x_{t-1}^{(i)},l_{t-1}),
\end{align}
\vspace{-0.5cm}
\begin{align}\label{16}
   &\delta_{t}^{(i)} = \frac{1}{\eta^{(i)}} \cdot \sum\limits_{j=1}^{z} p(o_{t}|\hat{m}_{t-1}^{(i)},x_{j}) \cdot p(x_{j}|x_{t-1}^{(i)},l_{t-1}) \nonumber \\
   & \cdot (x_{j}-\mu_{t}^{(i)})(x_{j}-\mu_{t}^{(i)})^{T}.
\end{align}
\vspace{-0.5cm}
\par
Thus, the x and y coordinates of each position \emph{$(x_{t}^{(i)},y_{t}^{(i)})$} of the \emph{i}-th iteration is derived from the Gaussian distribution \emph{$\mathcal{N}(\mu_{t}^{(i)},\delta_{t}^{(i)})$}. Thus, the average coordinate approximation is calculated as
\vspace{-0.2cm}
\begin{align}\label{17}
  x_{t} = \frac{1}{I} \sum\limits_{i=1}^{I}x_{t}^{(i)}, y_{t} = \frac{1}{I} \sum\limits_{i=1}^{I}y_{t}^{(i)}.
\end{align}
\vspace{-0.4cm}
\par
\noindent
where the \emph{I} denotes the number of the particles. Note that each particle is responsible for iterating once, The mentioned above is to method for localization, while the mapping work needs to be completed according to the Bresenham algorithm \cite{IEEEhowto:Bresenham}.

\begin{remark}\label{remark 1}
  For the basic principle of the Bresenham algorithm, Construct a set of virtual grid lines from the center of each row and column of pixels, and calculate the intersection of each vertical grid line of the straight line from the start point $S_{a_{n},b_{n}}$ to the final point $S_{a_{n+1},b_{n+1}}$, where $n \in \{1,2,\cdots,N\}$. The pixel of the column of pixels closest to this intersection point can be determined.
\end{remark}

Thus, since the method for map discretization, according to the \eqref{17}, the coordinates of the \emph{(a,b)}-th grid can be rewrriten as
\vspace{-0.4cm}
\begin{align}\label{18}
  x_{a,b} = \frac{1}{I} \sum\limits_{i=1}^{I}x_{a,b}^{(i)}, y_{a,b} = \frac{1}{I} \sum\limits_{i=1}^{I}y_{a,b}^{(i)},
\end{align}
\vspace{-0.4cm}
\par
\noindent
with the passible sub-map \emph{$m_{(a_{\hat{n}},b_{\hat{n}}) \rightarrow (a_{\hat{n}+1},b_{\hat{n}+1})}$}
\vspace{-0.3cm}
\begin{align}\label{19}
  m_{(a_{\hat{n}},b_{\hat{n}}) \rightarrow (a_{\hat{n}+1},b_{\hat{n}+1})} = p(\hat{m}|x_{1:\frac{2x_{max}}{\delta V}},o_{1:\frac{2x_{max}}{\delta V}}),
\end{align}
\vspace{-0.4cm}
\par
\noindent
where the \emph{$\hat{n} \in \{1,2,\cdots,\hat{N}\}$} denotes the number of the passible grids. Then, we assume the time consumed by SLAM computation at each grid is negligible. To evaluate the quality of SLAM, root mean square error (RMSE) can be utilized to count the absolute trajectory error (ATE) of robots reaching each grid, which can be expressed as
\vspace{-0.3cm}
\begin{align}\label{20}
  MSE(\pmb{X},\tilde{\pmb{x}}) = \frac{1}{Q_{0}}\sum\limits_{q=1}^{Q_{0}}(x_{a,b}^{(q)}-\tilde{x}_{a,b}^{(q)})^2,
\end{align}
\vspace{-0.4cm}
\par
\noindent
where \emph{$x_{a,b}^{(q)}$}, \emph{$\tilde{x}_{a,b}^{(q)}$}, \emph{$\pmb{X}$} and \emph{$\tilde{\pmb{x}}$} denote the observation data for the \emph{q}-th measurement at the \emph{(a,b)}-th grid, real data for the \emph{q}-th measurement at the \emph{(a,b)}-th grid, the set of observation data and the set of real data, respectively. According to the L-SLAM algorithm, the \emph{$\pmb{M}$} can be obtained. Additionally, global geographic map recovery (GGMR) algorithm can assist the positions of the robot teetotally shroud the whole environment, while all sub-maps can be established by L-SLAM algorithm. The GGMR algorithm is illustrated in the next subsection.

\subsubsection{Global geographic map recovery algorithm for sub-maps connection}
As the geographic map are discretized\footnote{In this paper, we do not consider improving the performance of SLAM algorithm.}, all the accessible grids are expanded in searching for an complete coverage path. Thus, a search algorithm must constantly make a decision about which grid to explore next. If it expands grids which obviously cannot be accessible, it loses the effort. On the other hand, if it ignores grids that can be admissible, it will fail to reconstruct the radio map for the indoor environment. An efficient global geographic map recovery algorithm for sub-maps connection obviously needs to evaluate available grids to determine which grid is accessible. Additionally, it is essential to avoid path duplication and shorten the moving distance during the robot traversal procedure. The robot shall turn 90 degrees or 180 degrees when it encounters an impassable region ahead. In the whole path search procedure, finding the optimal path between grids obeys the A* algorithm.
\par
As mentioned above, when building a geographic map and channel power gain map, the size of the grid can be ignored. However, the grid size is the main influencing factor of GGMR since the positions of the robot are determined by GGMR. Suppose \emph{$l_{0}$} as the length of each grid and the shape of each grid is square. On account of executing the GGMR algorithm in an unknown environment is messy, the problem can be concise to which rough environmental boundary is stipulated beforehand. Accordingly, the SLAM algorithm is iteratively tested in the same physical environment. The calculated MSE based on the equation \eqref{20} is utilized to pre-define the rough boundary. The proposed GGMR algorithm for exploring the whole environment is summarized in \textbf{Algorithm~\ref{GGMR}}. 
\begin{algorithm}[!t]
  \caption{GGMR algorithm for the whole environment}
  \label{GGMR}
  \begin{algorithmic}[1]
  \Require ~~\\
  Environment theoretical size \emph{$2x_{max} \times 2y_{max}$}, obstacle theoretical size set \emph{$\{l_{o}* w_{o}\}$}, Wall theoretical size \emph{$l_{w}*w_{w}$}, numbers of walls \emph{$N_{1}$}, number of obstacles \emph{$N_{2}$}, \emph{MSE} for SLAM algorithm.
  \Ensure Values of all grids and robot trajectory.
  \State \textbf{Initialize:} 
  \State Size (\emph{$2x_{max}$}-\emph{MSE})(\emph{$2y_{max}$}-\emph{MSE}) for environment, move direction set \emph{$\pmb{D}$}, map resolution \emph{$\delta$}, expansion radius \emph{$R_{e}$}, positive direction is positive half x-axis.
  \State Randomly select start position \emph{$S_{0}$} and final position \emph{$S_{f}$}.
  \State Current position is pre-defined as \emph{$S_{0}$}.
  \State The moving direction is the positive direction first, followed by the negative direction.
  \Repeat
  \State Explore the adjacent grid in the same direction as the movement at current position.
  \If {the adjacent grid = 0.5 or 1}
    \State Explore the three grids above at current grid.
    \If {There is one grid = 0}
      \State Select the move direction d from \emph{$\pmb{D}$}.
      \State Next position = current position + d;
    \ElsIf {the value of all three grids = 0.5 or 1}
      \State New start position needs to be defined.
      \State Go to step 5;
    \EndIf
  \ElsIf{the adjacent grid = 0}
    \State Select the move direction d from \emph{$\pmb{D}$}.
    \State Next position = Current position + d;
  \EndIf
  \Until The value of all grids are 0.5 or 1.
  \end{algorithmic}
\end{algorithm}

\vspace{-0.4cm}
\subsection{Radio map construction algorithm}
Since the geographic map is constructed, the radio map, namely, the channel power gain map can be simultaneously obtained,  where the channel power gain map is constructed by exploiting the information of exact channel propagations. Let \emph{$\pmb{F} \in \mathbb{R}^{X \times Y}$} denote the channel power gain map, the path loss can be patritioned into non-light-of-sight (NLoS) and light-of-sight (LoS). Thus, the expected effective channel power gain at \emph{(a,b)}-th grid is given by
\vspace{-0.2cm}
\begin{align}\label{21.1}
  &F_{a,b} = \mathbb{E}[\pmb{L}_{a,b}] = \left\{
    \begin{array}{lr}
      L_{{\rm LoS}}, {\rm\ AP-robot\ link\ is\ unblocked},&  \\
      L_{{\rm NLoS}}, {\rm\ otherwise},
    \end{array}
  \right.
\end{align}
\vspace{-0.4cm}
\par
\noindent
where the \emph{$L_{{\rm LoS}}$} and \emph{$L_{{\rm NLoS}}$} denote the path loss between the AP and \emph{(a,b)}-th grid with LoS and NLos cases, respectively. Note that for simplicity, for SLARM, no interference except for noise power \emph{$\sigma^{2}$} is introduced into the transmitted signal. So far, the channel power gain map in the unknown environment is obtained according to the SLARM framework. 

\vspace{-0.3cm}
\section{Numerical results}
\begin{figure}[ht]
  \setlength{\belowcaptionskip}{-0.5cm}
  \centering  
  \includegraphics[height=1.8in,width=3in]{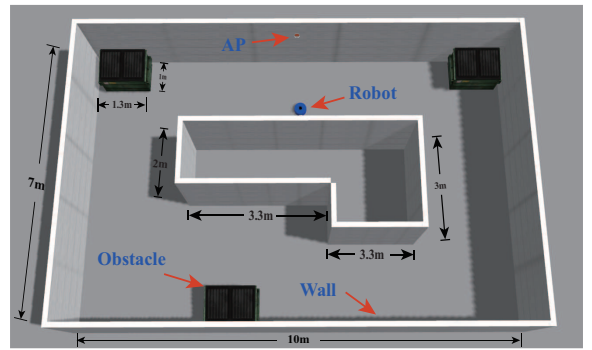}
  \caption{The geographic environment for simulation.}
  \label{room}
\end{figure}
The gazebo's model is leveraged to build a simulation environment, which provides high-fidelity physical simulation. When real robots and corresponding experimental environments are not available, it can genuinely mirror the real environment. As shown in Fig.~\ref{room}, the environment has two layers of walls and three regular cuboid obstacles. The rest of the ground between the two layers of walls is the possibly accessible regions for the robot. Also, the designated size was demonstrated, where the width, length, and ceiling height of the room are 10m, 7m, and 3m. Additionally, the obstacles are with a size of 1.3m $\times$ 1m $\times$ 1.5m, and the internal wall is with a total outer circumference of 21m, the thickness of 0.5m, and height of 1.8m. The AP and is deployed at (0,3.5,2.5). The path loss \emph{$L_{a,b}$} demonstrated in equation \eqref{21.1} according to report for the InF-SH scenario in \cite{IEEEhowto:3GPP-TR-38.901}, which can be expressed as
\vspace{-0.2cm}
\begin{align}\label{21.2}
  &F_{a,b} = \mathbb{E}[\pmb{L}_{a,b}] = \nonumber \\
  & \left\{
    \begin{array}{lr}
      L_{{\rm LoS}}=31.84+21.5{\rm log}_{10}(d_{a,b})+19{\rm log}_{10}(f_{c}),&  \\
      L_{{\rm NLoS}}=\max\{L_{LoS},32.4+23{\rm log}_{10}(d_{a,b})+20{\rm log}_{10}(f_{c})\},
    \end{array}
  \right.
\end{align}
\vspace{-0.45cm}

\vspace{-0.5cm}
\subsection{Radio map construction results}
\begin{figure*}[htbp] 
  \setlength{\belowcaptionskip}{-0.2cm}
  \setlength{\abovecaptionskip}{-0.15cm}
  \centering
  \subfigure[Theoretical power gain map, resolution = 0.05m.] {\label{Theoretical power gain map without IRS, resolution = 0.05m} \includegraphics[height=1.5in,width=2.2in]{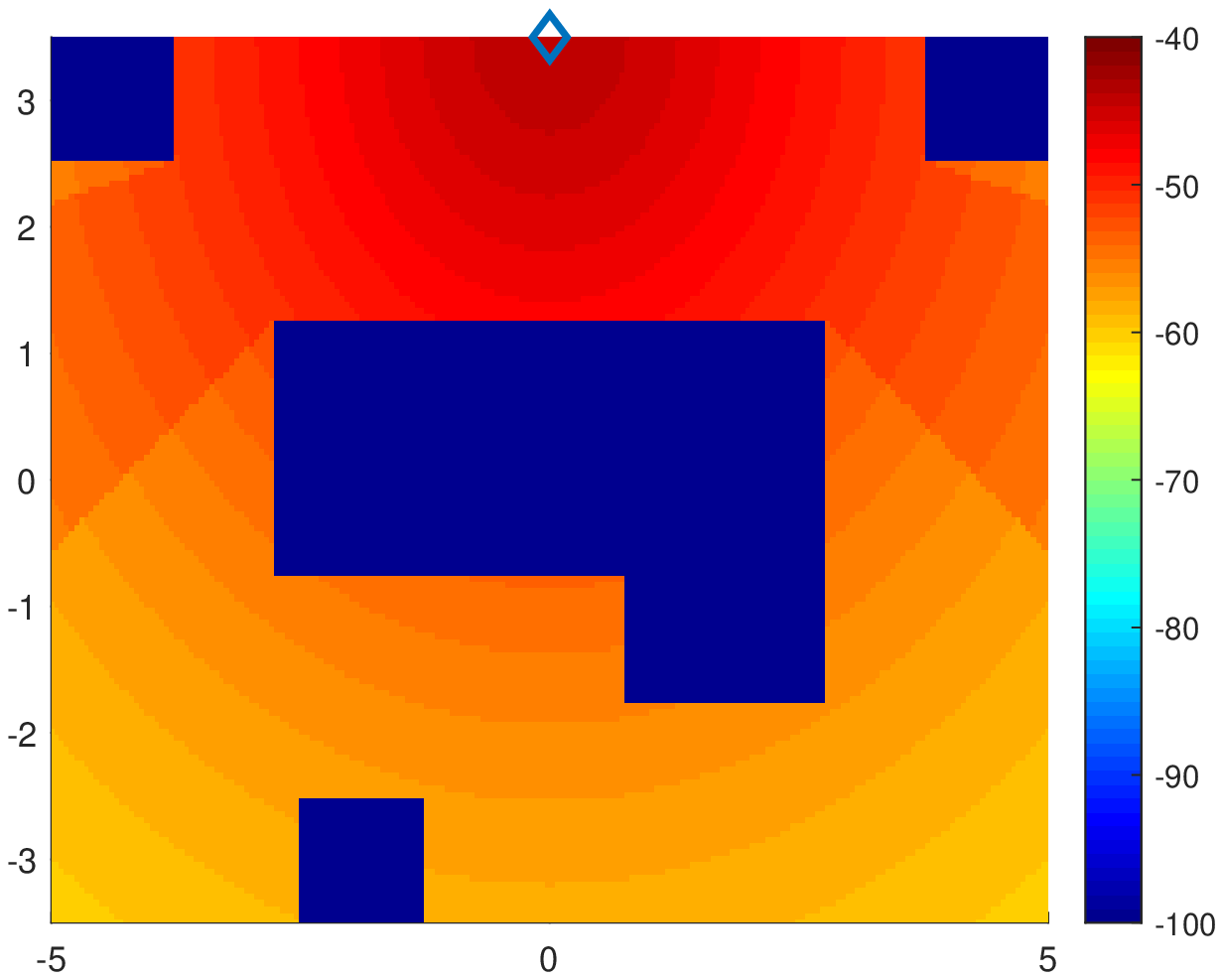}}\hspace{1mm}
  \subfigure[Theoretical power gain map, resolution = 0.1m.] {\label{Theoretical power gain map without IRS, resolution = 0.1} \includegraphics[height=1.5in,width=2.2in]{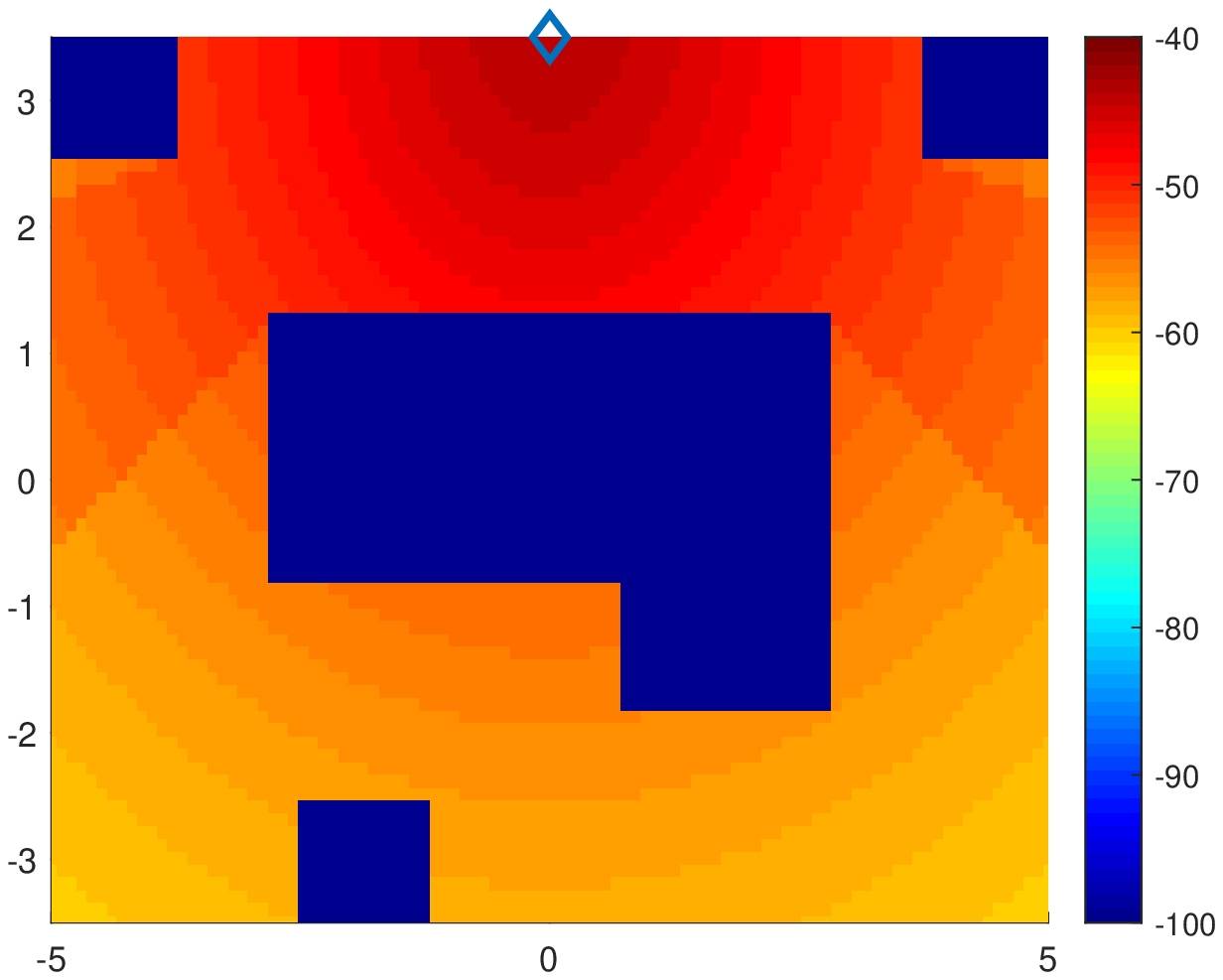}}\hspace{1mm}
  \subfigure[Theoretical power gain map, resolution = 0.25m.] {\label{Theoretical power gain map without IRS, resolution = 0.25} \includegraphics[height=1.5in,width=2.2in]{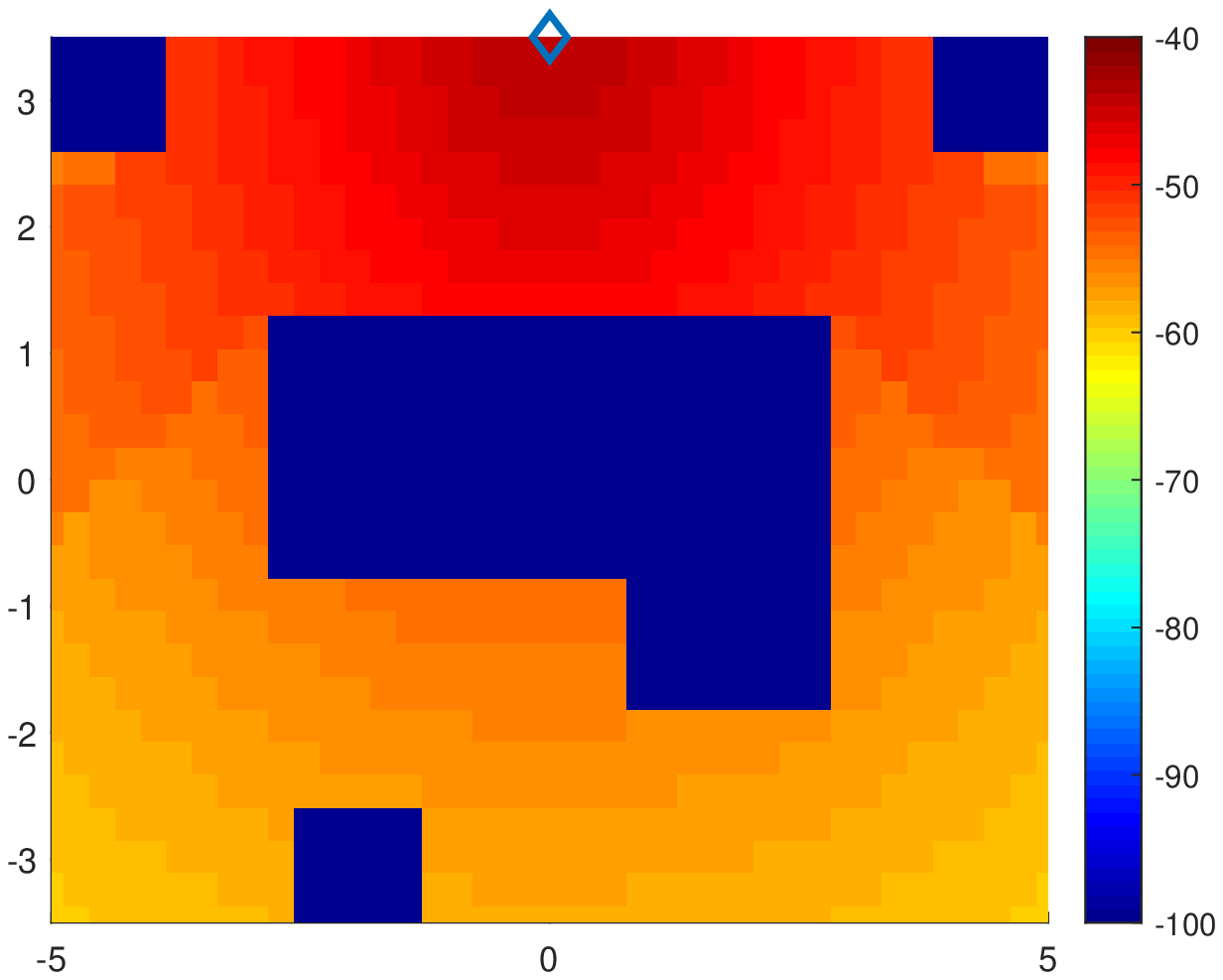}}\hspace{1mm} 
  \caption{Theoretical GGMR exploration and radio map construction.}
  \label{Theoretical GGMR exploration and radio map construction}
\end{figure*}
\begin{figure*}[htbp] 
  \setlength{\belowcaptionskip}{-0.2cm}
  \setlength{\abovecaptionskip}{-0.15cm}
  \centering
  \subfigure[Simulational power gain map, resolution = 0.05m.] {\label{Simulational power gain map without IRS, resolution = 0.05} \includegraphics[height=1.5in,width=2.2in]{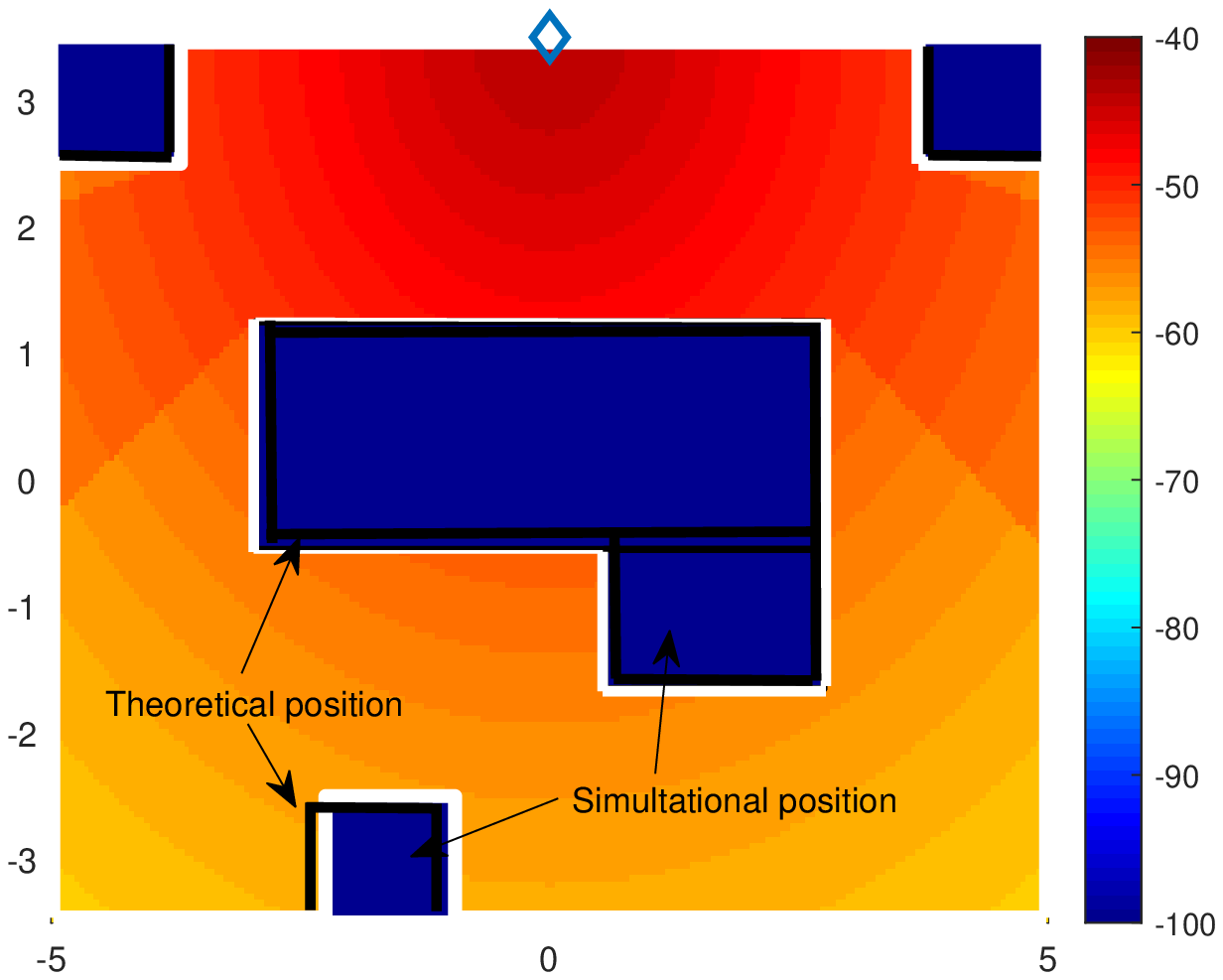}}\hspace{1mm}
  \subfigure[Simulational power gain map, resolution = 0.1m.] {\label{Simulational power gain map without IRS, resolution = 0.1} \includegraphics[height=1.5in,width=2.2in]{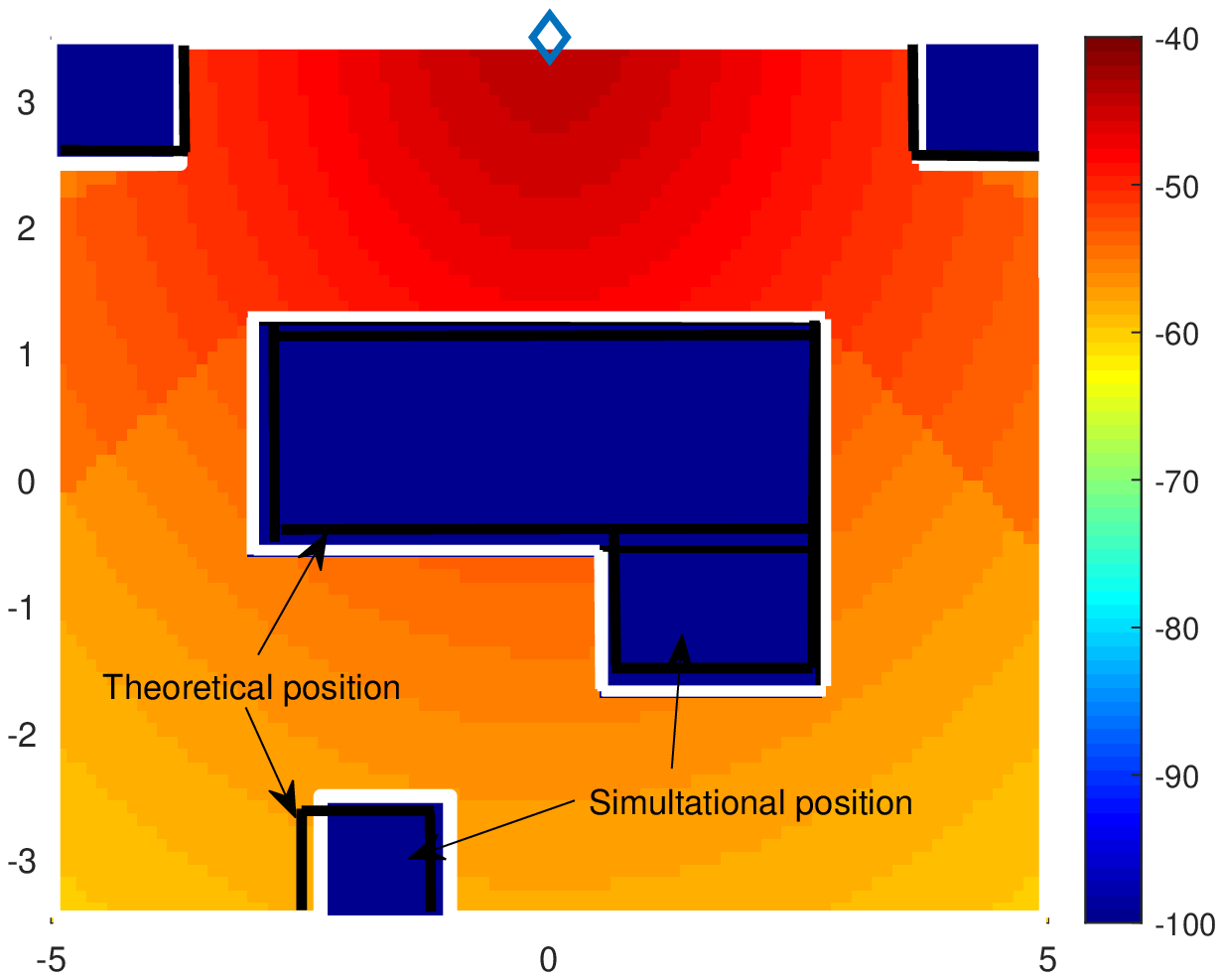}}\hspace{1mm}
  \subfigure[Simulational power gain map, resolution = 0.25m.] {\label{Simulational power gain map without IRS, resolution = 0.25} \includegraphics[height=1.5in,width=2.2in]{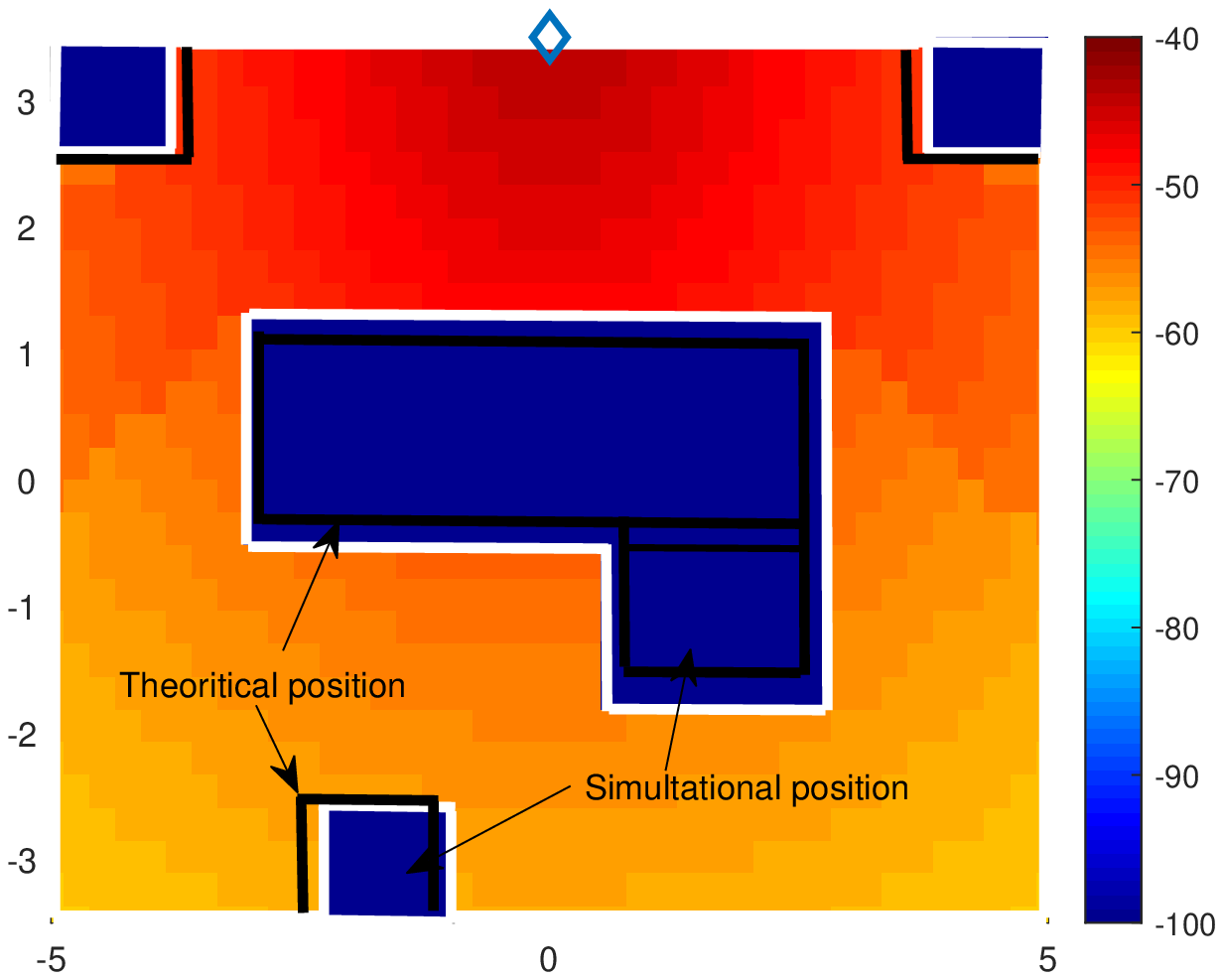}}\hspace{1mm}  
  \caption{Simulational GGMR exploration and radio map construction.}
  \label{Simulational GGMR exploration and radio map construction}
\end{figure*}
Our purpose is that the robot can simultaneously build the radio map and the geographic map. The gmapping algorithm and proposed GGMR algorithm are leveraged for geographic map construction since the environment needs to be explored. Owing to the errors of the SLAM algorithm and the GGMR algorithm, the position-dependent based channel power gain map in the exploration environment also has a bias. The two cases (theoretical and simulational) are considered, while the three different resolutions are enacted for comparison. The theoretical results are shown in Fig.~\ref{room}, where the size of the environment is pre-assumed as ideal, the channel power gain map turns blurry with the value of resolutions increased. Specifically, when the resolutions reach 0.25m, the size of some grids are not proper to approximate to the centers of grids. Thus, the channel power gain map is no longer efficient for future mission execution under such a resolution case, where the resource (like passable region space) is unexpected consumption. Additionally, the simulation results are shown in Fig.~\ref{Simulational GGMR exploration and radio map construction}, when the SLAM algorithm is employed, it can be obtained that due to the limitation of the algorithm performance, some burrs appear on the boundaries and the edges of obstacles. To resolve this problem, we pre-set an expansion radius near the boundary and obstacles, which is to prevent the robot from colliding with the boundary or obstacles. More imperatively, the positions of the obstacles deviate from the theoretical positions in the simulation results can bring the error to the channel power gain calculation, which also makes the large value of resolution is not permittable. Additionally, we found that according to the SLAM algorithm, the GGMR algorithm can be influenced to a certain extent. Then, in order to make the established radio map serve the future path planning, we have carried out edge detection and smoothing filtering on the image drawn in Fig.~\ref{Simulational GGMR exploration and radio map construction} without affecting the passable region since the boundaries and obstacles turn regular.

\vspace{-0.35cm}
\subsection{Accuracy for radio map construction}
\begin{figure}[htbp] 
  \setlength{\belowcaptionskip}{-0.4cm}
  \setlength{\abovecaptionskip}{-0.1cm}
  \centering  
  \includegraphics[height=2.2in,width=3in]{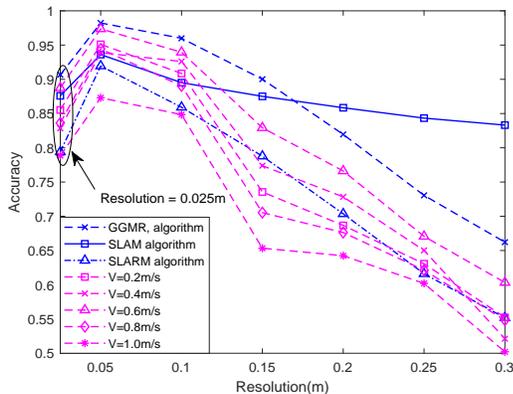}
  \caption{Resolution versus Algortihm accuracy and different robot speed.}
  \label{Resolution}
\end{figure}

In the process of building a radio map, the resolution is a decisive factor. The environment exploration time of the GGMR algorithm slows down gradually as the resolution decreases. It is worth noting that SLARM algorithm time is the sum of GGMR exploration time and radio map measurement time, where the measurement time can be ignored. It is announced that the completion time increases as the value of resolution increases, which is in line with our expectations. But it does not prove that the accuracy of the SLARM algorithm has a similar relationship with the size of the resolution, which should be further discussed. Thus, we define the velocity of the robot \emph{v=0.6m/s} and resolution as 0.02 to test it. It should be emphasized that the accuracy rate is determined by the MSE of the SLAM algorithm, the GGMR algorithm coverage grid rate and the robot speed. As shown in Fig.~\ref{Resolution}, under this definition, the accuracy is lower than when the resolution is 0.05. Additionally, we changed the velocity of the robot, as shown in Fig.~\ref{Resolution}, we can see that when the speed of the robot reaches 1m/s, SLARM has the highest accuracy 91.95\%. As for the SLAM algorithm, because this article did not improve and discuss the algorithm, and from the results, the change of MSE is consistent with the resolution. In this article, the accuracy of SLARM focuses on the performance of the GGMR algorithm and the speed of the robot.

\vspace{-0.4cm}
\subsection{Performance for SLARM algorithm}
As mentioned above, the performance of the SLARM occupies a pivotal position in the entire radio map establishment process. Furthermore, we analyze the performance of the GGMR algorithm. The GGMR algorithm proposed in this paper, let the robot traverses the whole map to explore its space complexity and time complexity by generating grids with different resolution values in the map. Its space complexity is mainly determined by map data, and its space complexity is \emph{$O(N^{2})$}. Time complexity needs to consider the number of comparisons in the exploration process. In this algorithm, in the best case, the robot needs to compare the three grids above its 8-connected area 3 times (3 cycles) for each grid and find the final position from the first position. In the worst case, the robot not only needs to perform these 3 comparisons (three-fold cycle failure), the robot cannot find the final position at one time, but also needs to perform \emph{$N$} position searches. The log\emph{$N$} level is divided and conquered under the final level of the cycle, and each level needs to be merged and searched \emph{N} times. In summary, it can be seen that the best time complexity is \emph{$O(N^{3})$}, and the worst time complexity is \emph{$O(N(logN)^{N})$}.

\vspace{-0.6cm}
\section{Conclusion}
A SLARM framework was proposed to comply with the construction of the radio map in the unknown environment for communication-aware connected robots. In this framework, combing the SLAM algorithm with the features of the wireless network, simultaneously geographic map and radio map construction has been implemented. A novel global geographic map recovery algorithm was responsible for assuring all the sub-maps can be connected. Numerical results showed that the resolutions of the SLAM algorithm and GGMR algorithm play a significant role in possible communication area utilization. The accuracy reaches 91.95\% when the resolution is pre-defined as 0.05m.

\ifCLASSOPTIONcaptionsoff
  \newpage
\fi

\end{document}